\def\year{2017}\relax
\documentclass[letterpaper]{article}
\usepackage{aaai17}
\usepackage{times}
\usepackage{helvet}
\usepackage{courier}
\usepackage{float}
\usepackage{graphicx} % more modern
\usepackage{subfigure}
\usepackage{amsmath}
\usepackage{algorithm}
\usepackage{algorithmic}
\usepackage{multicol}
\usepackage{multirow}
\begin{document}
\title{ Dual Teaching: A Practical Semi-supervised Wrapper Method}
\author{Fuqaing LIU$^{1}$ Chenwei DENG$^{1}$ Fukun BI$^{2}$ Yiding YANG$^{1}$\\
$^{1}$Department of Information and Electronic, Beijing Institute of Technology, Beijing\\
$^2$College of Information Engineering, North China University of Technology, Beijing\\
fqliu@outlook.com,   cwdeng@bit.edu.cn,   bifukun@ncut.edu.cn,   yangydi9g3@bit.edu.cn
}
%\author{AAAI Press\\
%Department \\
%Address\\
%Email
%}
\maketitle

\begin{abstract}
Semi-supervised wrapper methods are concerned with building effective supervised classifiers from partially labeled data. Though previous works have succeeded in some fields, it is still difficult to apply semi-supervised wrapper methods to practice because the assumptions those methods rely on tend to be unrealistic in practice. For practical use, this paper proposes a novel semi-supervised wrapper method, Dual Teaching, whose assumptions are easy to set up. Dual Teaching adopts two external classifiers to estimate the false positives and false negatives of the base learner. Only if the recall of every external classifier is greater than zero and the sum of the precision is greater than one, Dual Teaching will train a base learner from partially labeled data as effectively as the fully-labeled-data-trained classifier. The effectiveness of Dual Teaching is proved in both theory and practice.
\end{abstract}

\section{Intruduction}
Semi-supervised learning (SSL) is concerned with utilizing some labeled data as well as amounts of unlabeled data to train better classifiers and regressors~\cite{Balcan2010A,zhu2009introduction}. Some SSL methods are called "Semi-supervised wrapper methods"~\cite{semisupervisedsurvey} and they procedure external "wrappers" around the base learner, a supervised learning model. The purpose of semi-supervised wrapper methods is to train a supervised model from partially labeled data as effectively as the same model trained from fully labeled data. Compared with other semi-supervised methods, semi-supervised wrapper methods can use almost any user-specified supervised learner as a subroutine, which allows us to take advantage of advances in supervised learning without deriving new algorithms.

%However, it is difficult to apply semi-supervised wrapper methods in real-world learning cases. Previous wrapper methods do not work in many practical tasks and even worse they might be outperformed by simple supervised learner trained from only labeled data. Fig~\ref{fig:negative examples} shows two examples that the performance of semi-supervised wrapper methods grows worse when using unlabeled data.

%\begin{figure}[htb]\centering
%\vskip 0.1in
%\subfigure[]{\includegraphics[width=0.24\textwidth]{self-training.jpg}}
%\subfigure[]{\includegraphics[width=0.24\textwidth]{co-training.jpg}}\\
%%\subfigure[]{\includegraphics[width=0.33\textwidth]{c2.jpg}}
%\caption{Two examples show that previous semi-supervised wrapper methods might train the base learner worse when using unlabeled data. 10\% data are labeled in the two examples. In the first generation, the base learners are trained from the labeled data, and in the following generations, unlabeled data are also used in the training procedure. (a) shows the performance of self-training with logistic regression as the base learner. (b) shows the performance of co-training with logistic regression and SVM as the base learners.}
%\label{fig:negative examples}
%\vskip -0.1in
%\end{figure}

However, it is difficult to apply semi-supervised wrapper methods in practical learning cases~\cite{umvp,Blum2000Combining,Singh2008Unlabeled}. Though most of previous wrapper methods propose theoretical analysis to prove their effectiveness, there are still many gaps between realities and theories~\cite{zhousemi,Loog2015Contrastive,Sokolovska2008The}. In real-world learning scenarios, assumptions those semi-supervised wrapper methods rely on might be untenable, which makes those methods ineffective. For instance, self-training~\cite{selftraining2006Effective} assumes that predictions of the base learner tend to be correct in early generation otherwise it would grow worse when using the unlabeled data. However for a base learner initialized from a few labeled data, making good predictions on amounts of unlabeled data is a tough job. More related semi-supervised wrapper methods including Co-Training~\cite{Blum2000Combining} and Cluster-Then-Label-Methods~\cite{clusterthenlabel} as well as the reason why they are hard to set up in practice are discussed in Related Work~\ref{sec:related work}.

For practical use, this paper proposes a novel semi-supervised wrapper method named "Dual Teaching" whose assumptions are easy to set up in practical learning tasks. Dual Teaching only adopts two external classifiers outside the base learner. One assumption of Dual Teaching is that the recall of each auxiliary classifier is greater than zero. The other is that the sum of precisions is greater than one. Compared with assumptions of previous works, these assumptions are much easier to set up in practice. The two assumptions themselves are so simple that they can be matched even by two weak classifiers. In addition, the assumptions have no relationship with the base learner so that Dual Teaching can choose a proper supervised learner as the base learner according to the practical data.

The above easy assumptions originate from a novel framework. Different from previous works, auxiliary wrappers in Dual Teaching estimate the error of the base learner rather than directly estimate the label of unlabeled data. One external classifier named "FP Teacher" picks out false positives from the positive outputs of the base learner. The other one named "FN Teacher" picks out false negatives from the negative outputs. To begin with, the base learner is zero-initialized. Then an effective learner would be trained by the following recursive processing. In every generation, the unlabeled data are classified by the base learner firstly; secondly the two teachers are tested on a validation set composed by part of labeled data and if the teachers do not match the assumptions, they will be tuned based on the rest labeled data until their performance on validation set matches the assumptions; thirdly the two teachers point out the unlabeled examples mis-classified by the base learner and then the base learner is re-trained using all previous mis-classified examples with their corrected labels. The base learner would grow better in every generation if the assumptions are valid.

%Contributions of this paper are summarized as follows. First, this paper designs a novel framework (Dual Teaching) that adopts two external classifiers to estimate the errors of the base learner in every iteration. Second, this paper theoretically analyzes the effectiveness of Dual Teaching. When the recalls of the two teachers in Dual Teaching are greater than zero respectively and the sum of precision is greater than one, Dual Teaching would train a supervised model from partially labeled data as effectively as the same model trained form fully labeled data. The above assumptions are easy to be set up in practice. Third, the efficiency of Dual Teaching is analyzed in theory and several ways on improving the efficiency are proposed. Fourth, this paper presents an extensive evaluation of the state-of-the-art methods on benchmark data, where the proposed method achieved much better performance than previous semi-supervised wrapper methods. By Dual Teaching, supervised models trained from 20\% labeled data and 80\% unlabeled data are almost as effective as the fully-labeled-data-trained models.

The rest of the paper is organized as follows. We first review related semi-supervised wrapper methods. Next, the proposed method is detailed. Then we analyzes the effectiveness of Dual Teaching in theory. This is followed by a number of experiments to show the effectiveness in practice. This paper finishes with conclusions and suggestions for future research.

\section{Related Work}\label{sec:related work}
This section reviews a series of semi-supervised wrapper methods including Self-Training, Co-Training and Cluster-then-label-methods. Differences between Dual Teaching and the above methods are also discussed.

\textbf{Self-Training}~\cite{Mcclosky2006Effective} is characterized by the strategy that the learner uses its own predictions to teach itself. It starts by training a base learner from some labeled data and then evaluates the learner on the unlabeled data. Examples along with the labels predicted by the base learner are added to the training set and then the classifier is re-trained. Because the early mistakes made by the base learner would be reinforced by inputting the false labels to the early training set, self-training assumes that predictions of the base learner tend to be correct. However, it is unrealistic in some practical scenarios that a supervised model trained form a few labeled could classify amounts of unlabeled data successfully.

\textbf{Co-Training}~\cite{Balcan2004Co} is a wrapper method that works with two classifiers. The learning is initialized by training two classifiers from two separate feature sets of the labeled data. Then, one classifier make predictions on unlabeled data and then the data with their predicted labels are fed back for re-training another classifier. The two classifiers work as the above step alternately. The high performance of Co-Training relies on two independent feature-spaces. In addition, each classifier should make good predictions on unlabeled data in early generation. The above two assumptions tend to be violated in practice. First, two independent feature-spaces are difficult to be available because the diffusion of the real-world data is complicated. Second, it is difficult for the two classifiers to make good predictions on unlabeled data at the very start when they are trained from a few labeled data.

\textbf{Cluster-then-label-Methods}~\cite{clusterthenlabel} start by identifying clusters by unsupervised clustering algorithms and unlabeled data. Then supervised classifiers are learned from the labeled examples in each cluster. Each cluster should learn a classifier and if there is no labeled example falls into the cluster, the classifier would be trained from all labeled data. After that, the unlabeled examples in each cluster are labeled by the corresponding classifier. Cluster-then-label-methods will work effectively only if one cluster contains examples belong to a single class only. But in practice, one cluster usually contains examples from more than one class.

Dual Teaching is very different from previous works. Dual Teaching makes assumptions on its wrappers rather than the base learner or the data. Accordingly, the wrappers can be adjusted to match the assumptions in the process of Dual Teaching. The special assumptions are derived from a novel framework. Dual Teaching tries to estimate the error of the base learner, while previous works try to directly estimate the labels of unlabeled data. Thanks to this strategy, Dual Teaching can provide theoretical analysis for general cases with arbitrary inner base learners.

\section{Description of Dual Teaching}
\label{sec:dual teaching}
We focus on binary classification. Suppose $x$ is an example from a feature-space $X$ and $y$ is its label from a label space $Y=\{1,-1\}$. Suppose $X_l$ is a labeled data set, $L_l$ is its corresponding label set and $X_u$ is an unlabeled data set, where $l$ and $u$ represent the number of labeled examples and that of unlabeled examples respectively. Assuming that $L_u$ is an in-existent label set that contains true labels of the unlabeled data, suppose $F:X\rightarrow Y$ be an supervised classifier learned from an assumed fully labeled data set $\{X_l, X_u\}$ with its label set $\{L_l, L_u\}$. Dual Teaching aims to train an effective classifier $f$ from a few labeled data $X_l$ with $L_l$ and amounts of unlabeled data $X_u$, where $l<<u$. The classifier $f$ named the "base learner" is a supervised classification model parameterized by $\Theta$. Dual Teaching try to train the base learner as effectively as the fully-labeled-data-trained classifier.

Dual Teaching contains four base steps: zero-initialization, classification, teacher testing \& tuning and error estimation \& re-training. Dual Teaching adopts two external classifiers named "FP Teacher" and "FN Teacher" to estimate the false positives and false negatives respectively from the outputs of base learner. The assumptions of Dual Teaching are that the recall of each teacher is greater than zero and the sum of precision is greater than one. When the assumptions are valid, Dual Teaching can train a effective base learner from partially labeled data.

\begin{figure}[ht]
\vskip 0.1in
\begin{center}
\centerline{\includegraphics[width=\columnwidth]{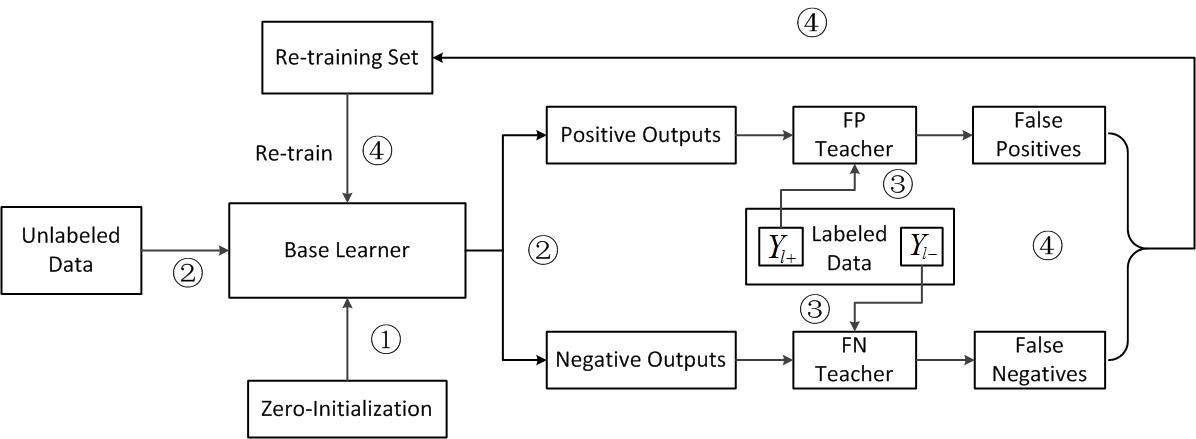}}
\caption{The framework of Dual Teaching.}
\label{fig:framework}
\end{center}
\vskip -0.1in
\end{figure}

\textbf{Step1: Zero-Initialization} For convenience, set the parameters of the base learner to zero, $\Theta^0=\textbf{0}$. In terms of Dual Teaching, how to initialize the base learner does not influence the performance of the base learner, which is different from Self-training and Co-training. Supposing $R$ represents the re-training data set for the base learner, it is an empty set initially, $R^0=\emptyset $. This step is shown as Fig~\ref{fig:framework} $\textcircled{1}$.

\begin{figure*}[htb]\centering
%\vskip 0.1in
\subfigure[precision/labeled]{\includegraphics[width=0.19\textwidth]{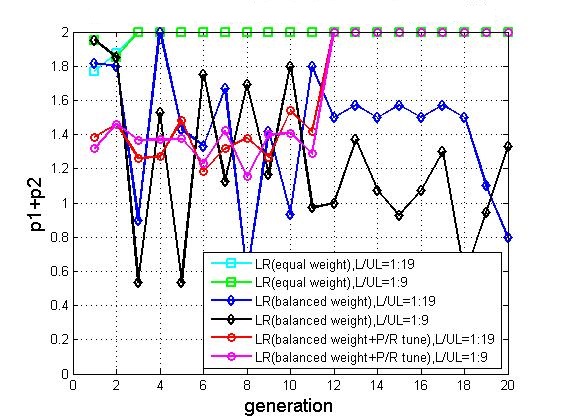}}
\subfigure[precision/unlabeled]{\includegraphics[width=0.19\textwidth]{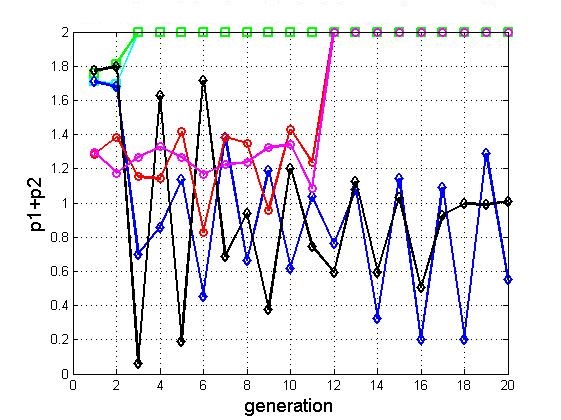}}
\subfigure[recall/labeled]{\includegraphics[width=0.19\textwidth]{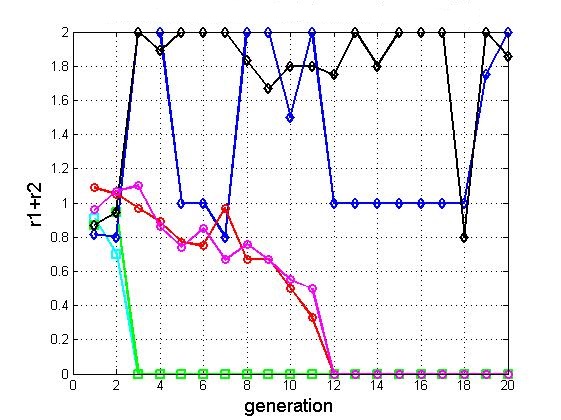}}
\subfigure[recall/unlabeled data]{\includegraphics[width=0.19\textwidth]{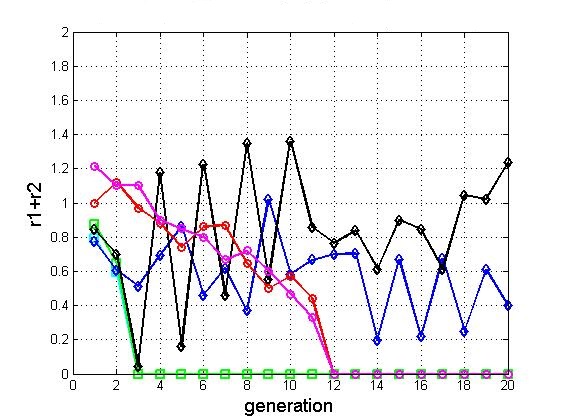}}
\subfigure[accuracy]{\includegraphics[width=0.19\textwidth]{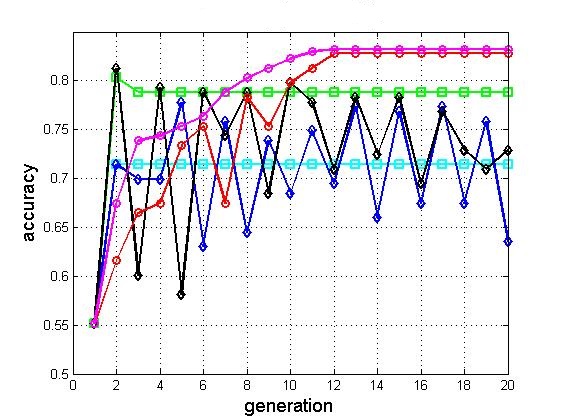}}
\caption{Performance of different strategies on constructing teachers. L/U in the legend represents the ratio of labeled data to unlabeled data in training set. (a-e) use a same legend in (a). The experiment is done on a dataset named heart scale.}
\label{fig:toy}
\vskip -0.1in
\end{figure*}

\textbf{Step2: Classification} In every generation, all unlabeled examples are classified by the base learner re-trained in last generation. $y^k=f(x,\Theta^{k-1})$ for $x \in X_u$ where $k$ represents the $k$th generation . This step is shown as Fig~\ref{fig:framework} $\textcircled{2}$.

\textbf{Step3: Teacher Testing \& Tuning} First, the auxiliary classifiers are trained from half of the labeled data which is referred to as $TX_l$. The rest labeled data are reserved as a validation set named $VX_l$. Then before estimating the error of the base learner, the teachers are tested on the validation set to make sure they match the assumptions. We estimate the true precision and recall of auxiliary classifiers by their recall and precision on validation set. If teachers do not match the assumptions, we will increase the precision of teachers by reducing their recall. Take logistic regression for instance, the threshold of discrimination, referred to as $T$ , is usually set to 0.5. But if we want to improve the precision, we will set $T$ greater than 0.5 and the recall will decrease accordingly.

This step, Teacher Testing \& Tuning, is shown as Fig~\ref{fig:framework} $\textcircled{3}$. The specific strategy is detailed as algorithm~\ref{pesudocode}. In the algorithm~\ref{pesudocode}, $FP$ and $FN$ represent the FP teacher and FN teacher respectively. $p1$ and $r1$ represent the precision and recall of FP teacher. $p2$ and $r2$ represent the precision and recall of FN teacher. %$T$ is the threshold of discrimination.
Allowing for a certain error, we empirically set the sum of precision should be greater than 1.1 rather than 1.

\begin{algorithm}[htb]
    \caption{Teaching Tuning}
    \label{pesudocode}
\begin{algorithmic}
    \STATE {\bfseries Input:} the validation set $VX_l$, the rest labeled data $TX_l$, the base learner $f$, the parameters $\Theta^k$
    \STATE {\bfseries Initialize}: $T$=0.1
    \STATE {\bfseries Data Division:}
    \STATE $V_{l+}=\{x\in VX_l|f(x,\Theta^k)==1\}$
    \STATE $V_{l-}=\{x\in VX_l|f(x,\Theta^k)==-1\}$
    \STATE $Y_{l+}=\{x\in TX_l|f(x,\Theta^k)==1\}$
    \STATE $Y_{l-}=\{x\in TX_l|f(x,\Theta^k)==-1\}$
    \STATE {\bfseries Teacher Training:}
    \STATE train $FP$ from $Y_{l+}$ with its true label
    \STATE train $FN$ from $Y_{l-}$ with its true label
    \REPEAT
        \STATE test $FP$ with $T$ on $V_{l+}$ and compute $p_{1}$,$r_{1}$
        \STATE test $FN$ with $T$ on $V_{l-}$ and compute $p_{2}$,$r_{2}$
        \STATE $T$ = $T$+0.05
        \UNTIL $(p1+p2>1.1\ and\ r1*r2>0)\ or\ (T>1)$
\end{algorithmic}
\end{algorithm}

As the base learner improves, it would be harder for teachers to correctly point out its mistakes because the proportion of mis-classified examples reduces. The two teachers gradually face a class imbalance problem as the base learner grows better. In practice, we construct each teacher by a balanced-weight logistic regression model~\cite{He2009Learning}. The threshold of discrimination is set to a small value initially for high recall. High recall means that teachers can pick out as many mis-classified examples as they could in step4: error estimation \& re-training.

The proposed strategy on constructing teachers is compared with other two strategies. One comparative strategy abbreviated to LR(equal weight) is to construct the teachers using equal-weight logistic regression. Another one abbreviated to LR(balanced weight)is to construct the teachers using balanced-weight logistic regression with a constant threshold. LR(balanced weight+L/R tuned) is short for the proposed strategy. The precision and recall of the base learner on both labeled set and unlabeled set are recorded in every generation. The performance of the base learner on unlabeled data is also recorded.

As Fig~\ref{fig:toy} shows, the recall of the base learner when using LR(equal weight) drops fast because the FP teacher and FN teacher can not find any mis-classified example in early generation. The accuracy of the base learner when using LR(balanced weight) shakes because the two teachers cannot match the assumptions in every generation. The base learner when using LR(balanced weight+L/R tuned) performs well because the teachers not only point out as many mis-classified examples as they can but also do not make too many mistakes (precision is always constant) as the base learner improves. Finally all of the outputs of two teachers are -1 which means the teachers think that the base learner makes no mistake. The performance of the base learner would be stable because the re-training set for the base learner does not change in next generation.

\textbf{Step4: Error Estimation \& Re-training} Dual Teaching adopts two external classifiers, FP Teacher and FN Teacher, to estimate the mis-classified examples in every generation. FP Teacher tries to pick out the false positives from the positive outputs of the base learner. FN Teacher tries to pick out the false negatives from the negative outputs. Then "add" these mis-classified examples with their estimated labels to the re-training set $R$. The "add" is defined as follows: if the mis-classified example already exists in re-training set $R$, just change its label; if the mis-classified example does not exit in the re-training set $R$, add the example with its corrected label to the re-training set. Finally, re-train the base learner from the updated re-training data set $R^k$. This step is shown as Fig~\ref{fig:framework} $\textcircled{4}$. Repeat steps 2-4 until the base learner is stable, then a strong classifier will be trained.

\section{Effectiveness Analysis}
\label{sec:effectiveness analysis}
To begin with, this section theoretically analyzes why Dual Teaching could train an effective classifier from partially labeled data. Then, we detail how to drive the assumptions Dual Teaching relies on. The assumptions are summarized as that the recall of each external classifier is greater than zero and the sum of precision is greater than one.

\subsection{Effectiveness}
\label{subsec:effectiveness}
This part theoretically derives that Dual Teaching could train an effective base learner from a few labeled data and amounts of unlabeled data just by two external classifiers.

Define $\overrightarrow{e}(k)$ as the error of the base classifier in kth generation,
\begin{equation} \label{error}
\overrightarrow{e}(k)=[\alpha(k), \beta(k)]^{T}
\end{equation}
%\end{center}
where $\alpha(k)$ represents false positives and $\beta(k)$ represents false negatives in the outputs of the base learner.

Define $P^{+}$,$R^{+}$ as the precision and the recall of FP Teacher respectively. Similarly, define $P^{-}$,$R^{-}$ as the precision and the recall of FN Teacher. Assume that if the mis-classified examples are pointed out and put into re-training set, the base learner will not make same mistakes in next generation. Assume that the base learner will classify the examples incorrectly if the examples along with wrong labels are contained in the re-training set. The recursive equations are shown as equation~(\ref{equ:recursive equations}).

\begin{equation}
\label{equ:recursive equations}
\left\{
\begin{aligned}
\alpha(k+1)=\alpha(k)-R^{+}\cdot\alpha(k)+\frac{1-P^{-}}{P^{-}}R^{-}\cdot\beta(k)\\
\beta(k+1)=\beta(k)-R^{-}\cdot\beta(k)+\frac{1-P^{+}}{P^{+}}R^{+}\cdot\alpha(k)
\end{aligned}
\right.
\end{equation}
where
\begin{itemize}
\item $R^{+}\cdot\alpha(k)$: the false positives that the FP Teacher pick out correctly.
\item $R^{-}\cdot\beta(k)$: the false negatives that the FN Teacher pick out correctly.
\item $\frac{1-P^{+}}{P^{+}}R^{+}\cdot\alpha(k)$: examples incorrectly classified to be false positives by FP Teacher.
\item $\frac{1-P^{-}}{P^{-}}R^{-}\cdot\beta(k)$: examples incorrectly classified to be false negatives by FN Teacher.
\end{itemize}
Reconstruct equation~(\ref{error}) and~(\ref{equ:recursive equations}),
%\begin{center}
\begin{equation}\label{equ:error}
\overrightarrow{e}(k+1)=M\overrightarrow{e}(k)
\end{equation}
%\end{center}
where $M=[\begin{array}{cc}
1-R^{+} & \frac{1-P^{-}}{P^{-}}R^{-}\\
\frac{1-P^{+}}{P^{+}}R^{+} & 1-R^{-}\\
\end{array}]$.

Based on the well founded theory of dynamic system~\cite{ROControl}, $\overrightarrow{e}(k)$ would converge to zero if all magnitudes of eigenvalues of the transformation matrix are smaller than one. The above analysis refers to the error analysis in Kalal's work~\cite{kalal2010pn,kalal2012tracking}.

Originally we assume that the base learner would not make the same mistakes as those picked out by two teachers in last generation, which is not always valid in practice. For instance, a classifier trained from fully-labeled data with their true labels would make mistakes on training set. Thus, the error of the base learner $f$ would converge to a certain level, a level as same as that of the fully-labeled-data-trained-classifier $F$.

\begin{figure}[htb]\centering
\vskip 0.1in
\subfigure[]{\includegraphics[width=0.23\textwidth]{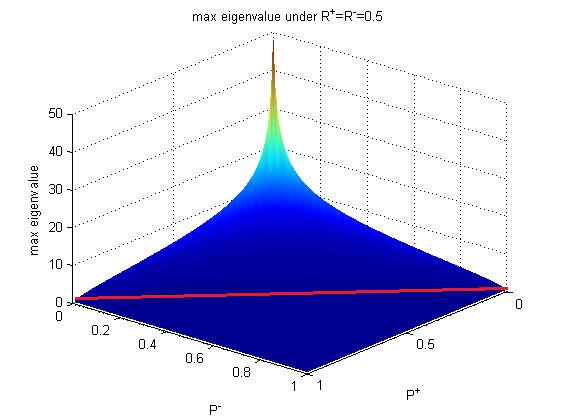}}
\subfigure[]{\includegraphics[width=0.23\textwidth]{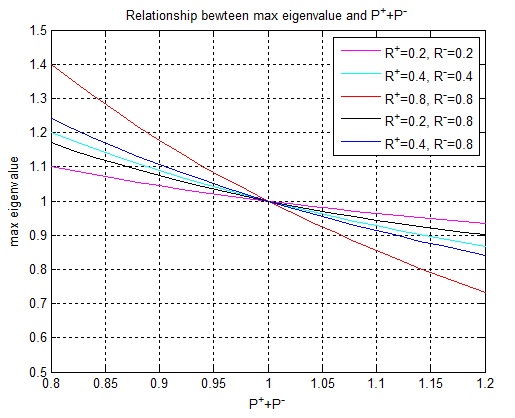}}\\
\caption{Simulations on $|\lambda|_{max}$. (a) shows the relationship between $|\lambda|_{max}$ and ($P^+$, $P^-$). %under the condition that $R^+=R^-=0.5$;
%(b) shows the relationship between $|\lambda|_{max}$ and ($P^++P^-$)
%In Fig~\ref{fig:similations}(a),
The red line is $P^++P^-=1$ and $|\lambda|_{max}$ in the red line is definitely equal to one. When $P^{+}+P^{-}>1$, $|\lambda|_{max}$ is definitely smaller than one. %Fig~\ref{fig:similations}
(b) shows that no matter what the recalls of the two classifiers are, the maximum magnitude of eigenvalue is definitely equal to one when $P^++P^-=1$. %The simulation results match equation~(\ref{equ:final conclusion}). %Only if the sum of the precisions of the two external classifiers (FN and FP Teachers) is greater than one, the magnitudes of the transformation matrix eigenvalues would be all smaller than one, and the error of the base learner would keep reducing.
}.
\label{fig:similations}
\vskip -0.2in
\end{figure}

\subsection{Assumption Derivation}
\label{subsec:assumptions}
This part details the assumptions Dual Teaching makes. Shown as equation~(\ref{equ:error}), the error continues reducing when the magnitudes of eigenvalues of the transformation matrix are smaller than 1. The eigenvalues $\lambda_1,\lambda_2$ are shown as equation~(\ref{equ:eigenvalues}).

\begin{equation}
\label{equ:eigenvalues}
\lambda_1,\lambda_2=\frac{(2-R^{-}-R^{+})\pm \sqrt{\bigtriangleup}}{2}
\end{equation}

\begin{equation}\label{equ:bigtri}
\bigtriangleup=(R^{+}-R^{-})^2+4 \frac{(1-P^+)(1-P^-)}{P^+ P^-}R^{+}R^{-}
\end{equation}
where $P^+,R^+,P^-,R^-\in [0,1]$ and $\bigtriangleup$ is always greater than or equal to 0.

So, the maximum absolute value of the two eigenvalues is shown as equation~(\ref{equ:max eigenvalue}).

\begin{equation}
\label{equ:max eigenvalue}
|\lambda|_{max}=\frac{(2-R^{-}-R^{+})+ \sqrt{\bigtriangleup}}{2}
\end{equation}
Assuming that $|\lambda|_{max}$ is smaller than 1, we can derive a conclusion as equation~(\ref{equ:conclusion}).
\begin{equation}
\label{equ:conclusion}
\sqrt{\bigtriangleup}<R^{-}+R^{+}
\end{equation}
Simplify equation~(\ref{equ:conclusion}) and we can get equation~(\ref{equ:condition}).
\begin{equation}
\label{equ:condition}
\frac{1-P^+-P^-}{P^+ P^-}R^+R^-<0
\end{equation}
Furthermore, the assumptions Dual Teaching makes are derived as equation~\ref{equ:final conclusion}.

\begin{equation}
\label{equ:final conclusion}
\left\{
\begin{aligned}
R^{+}R^{-}\neq0\\
P^{+}+P^{-}>1
\end{aligned}
\right.
\end{equation}

%Equation~(\ref{equ:final conclusion}) guides how to design the two external classifiers. When the recall of each classifier is greater than 0 and the sum of the precisions is greater than 1, the base learner will be trained effectively by Dual Teaching from partially labeled data.

%
%In Fig~\ref{fig:similations}(a), the red line is $P^++P^-=1$ and $|\lambda|_{max}$ in the red line is definitely equal to one. When $P^{+}+P^{-}>1$, $|\lambda|_{max}$ is definitely smaller than one. Fig~\ref{fig:similations}(b) shows that no matter what the recalls of the two classifiers are, the maximum magnitude of eigenvalue is definitely equal to one when $P^++P^-=1$. The simulation results match equation~(\ref{equ:final conclusion}). Only if the sum of the precisions of the two external classifiers (FN and FP Teachers) is greater than one, the magnitudes of the transformation matrix eigenvalues would be all smaller than one, and the error of the base learner would keep reducing.

\begin{figure*}[t]\centering
\vskip 0.1in
\subfigure[bci]{\includegraphics[width=0.22\textwidth]{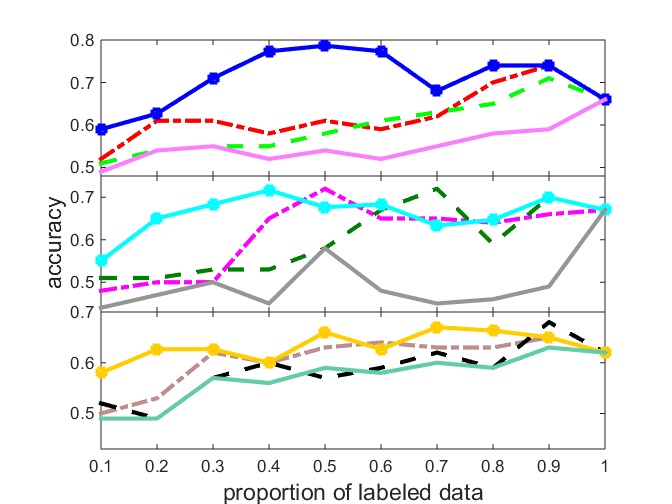}}
\subfigure[g241c]{\includegraphics[width=0.22\textwidth]{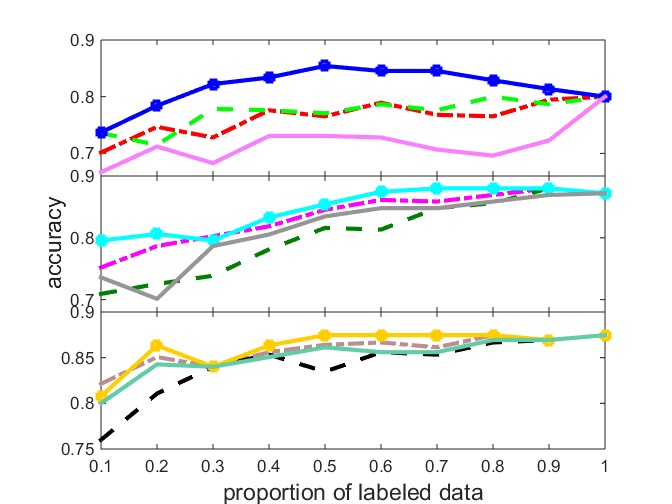}}
\subfigure[digit1]{\includegraphics[width=0.22\textwidth]{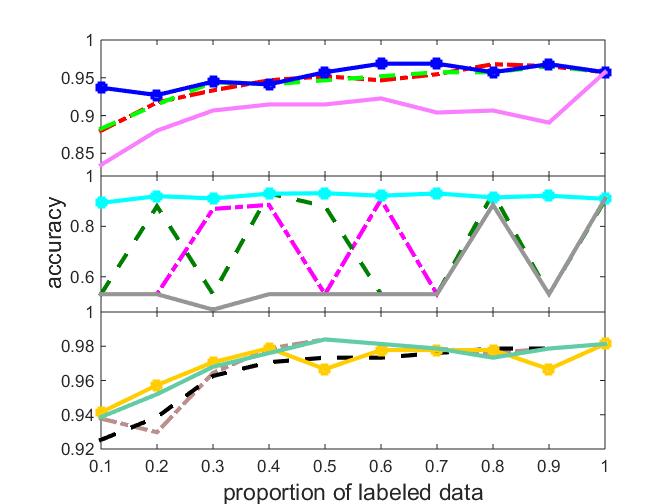}}
\subfigure[usps]{\includegraphics[width=0.22\textwidth]{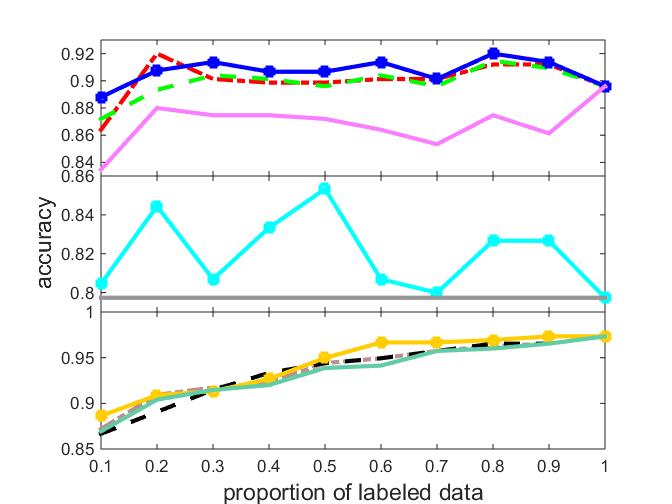}}\\
\subfigure[heart scale]{\includegraphics[width=0.22\textwidth]{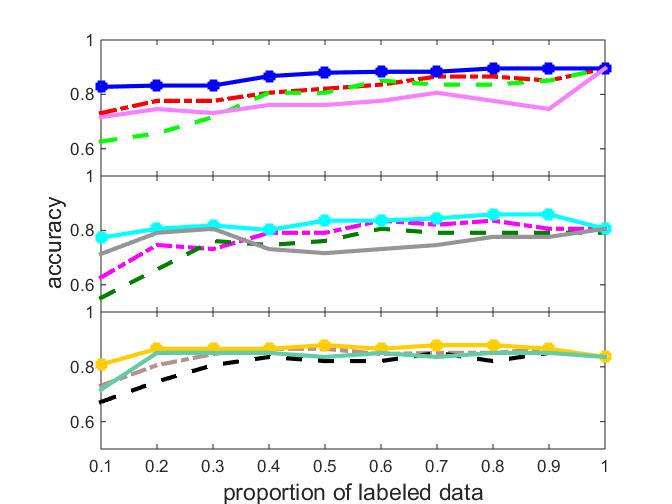}}
\subfigure[car evaluation]{\includegraphics[width=0.22\textwidth]{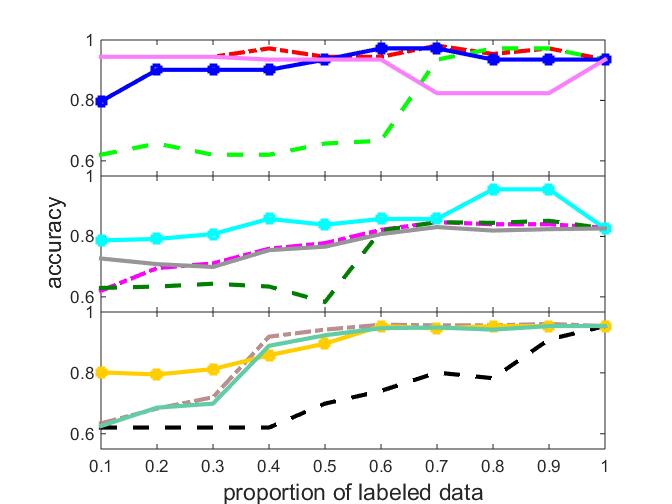}}
\subfigure[wine quality]{\includegraphics[width=0.22\textwidth]{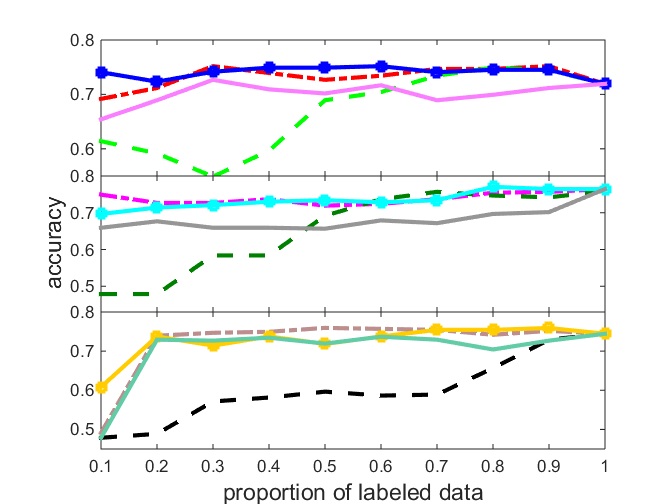}}
\subfigure[adult-a]{\includegraphics[width=0.22\textwidth]{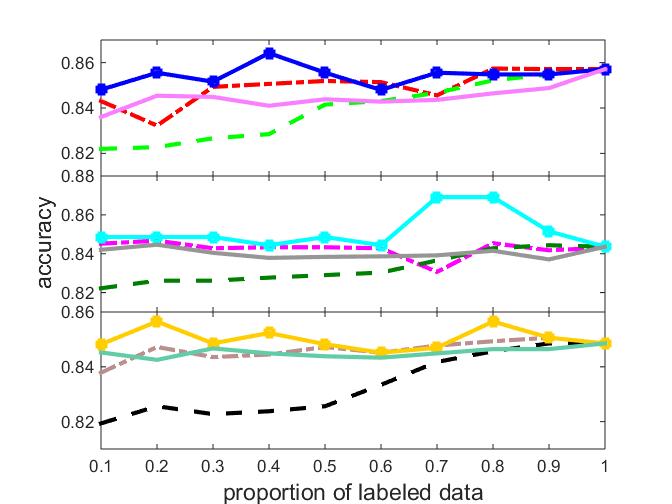}}\\
\subfigure[housing]{\includegraphics[width=0.22\textwidth]{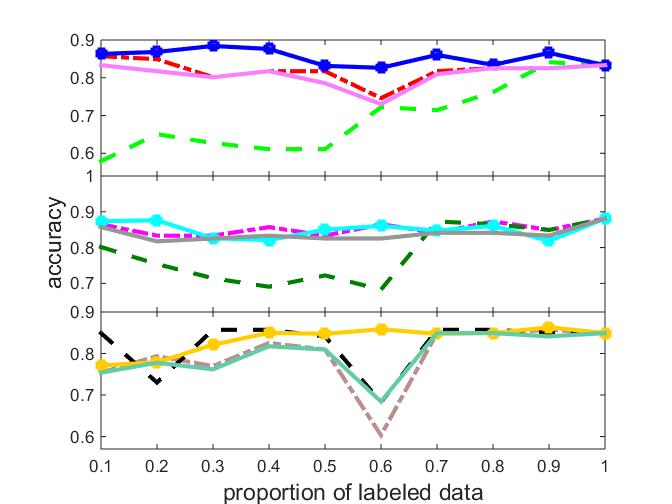}}
\subfigure[vehicle]{\includegraphics[width=0.22\textwidth]{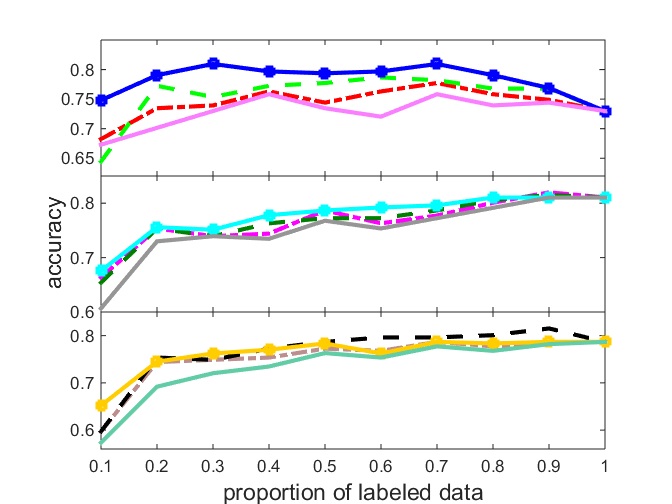}}
\subfigure[mnist1v7]{\includegraphics[width=0.22\textwidth]{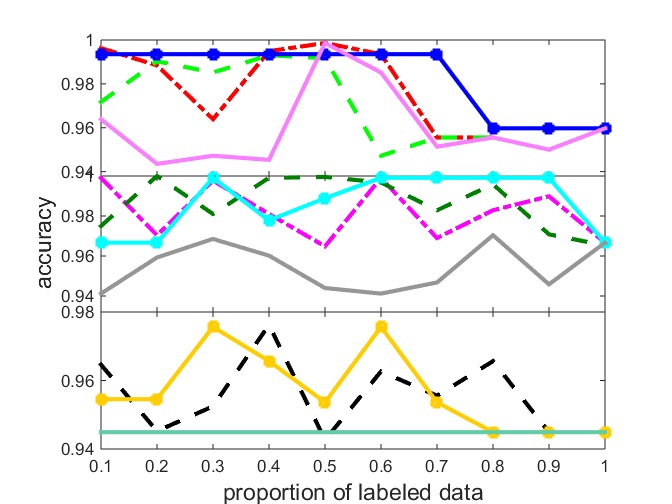}}
\subfigure[mnist4v9]{\includegraphics[width=0.22\textwidth]{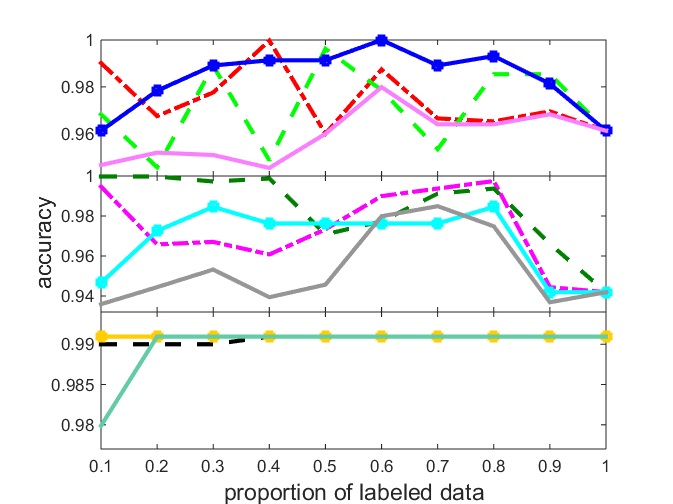}}\\
\subfigure{\includegraphics[width=0.6\textwidth]{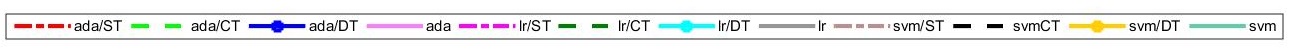}}\\
\vskip -0.1in
\caption{Performance of different semi-supervised wrapper methods with different base learners.}
\label{fig:exp}
\vskip 0.1in
\end{figure*}

Equation~(\ref{equ:final conclusion}) proposes the requirement on the two external classifiers. When the recall of each classifier is greater than 0 and the sum of the precisions is greater than 1, the base learner will be trained effectively by Dual Teaching from partially labeled data.

\section{Experiments}
This section proposes a series of evaluations of the proposed method and previous works on a broad range of tasks including 4 semi-supervised benchmark data sets\footnote{http://www.kyb.tuebingen.mpg.de/ssl-book/.}, \emph{bci}, \emph{g241c}, \emph{digit1}, \emph{usps}, and 6 UCI tasks\footnote{http://archive.ics.uci.edu/ml/datasets.html.}, \emph{heart scale}, \emph{car evaluation}, \emph{wine quality}, \emph{adult}, \emph{housing}, \emph{vehicle}, and 2 image classification problems~\cite{Lecun2010et}, \emph{mnist1v7}, \emph{mnist4v9}. The size of data ranges from 170 to more than 15000 and the dimensionality ranges from 6 to more than 600. The ratio of the number positive examples to that of negative examples ranges from 0.25 to 1.149. %Table~\ref{table:data sets} shows the characteristics of the data sets.
%
%\begin{table}
%\vskip -0.1in
%\begin{center}
%\caption{Characteristics of the Data Sets}
%\label{table:data sets}
%\begin{tabular}{l|ccc}
%\hline\noalign{\smallskip}
%Data Sets & Dim & pos/neg & Instance \\
%%\noalign{\smallskip}
%\hline
%\noalign{\smallskip}
%bci & 117 & 1 & 400\\
%g241c & 241 & 0.958 & 1500\\
%digit1 & 241 & 0.958 & 1500\\
%usps & 241 & 0.25 & 1500\\
%heart scale & 13 & 0.8 & 270\\
%car evaluation & 6 & 0.428 & 1728\\
%wine quality& 10 & 1.149 & 4898\\
%adult-a & 14 & 0.314 & 15345\\
%housing & 13 & 1.032& 506\\
%vehicle & 18 & 0.962 & 846\\
%mnist1vs7 & 652 & 1.080 &15170\\
%mnist4vs9 & 629 & 0.981 &13782\\
%\hline
%\end{tabular}
%\end{center}
%\vskip -0.1in
%\end{table}
%\setlength{\tabcolsep}{3pt}

Every data set is separated into two parts. 75\% data in each data set are reserved as training data. The testing data set is composed by the rest 25\% data. Not all data in the training data set are labeled. Features in each data set are normalized beforehand.

\begin{table*}[htb]
\vskip -0.1in
\begin{center}
\caption{Accuracy of svm and accuracy improvements of S3VM, S4VM and Dual Teaching on different numbers of labeled data}
\label{table:comparison}
\begin{tabular}{l|cccc|cccc|cccc}%|cccc}
\hline\noalign{\smallskip}
\multirow{2}{*}{Dataset} & \multicolumn{4}{|c|}{5\% labeled data} &  \multicolumn{4}{|c|}{10\% labeled data}& \multicolumn{4}{|c}{20\% labeled data}\\%&\multicolumn{4}{|c}{40\% labeled data}\\
& svm  &s3vm & s4vm & svm/DT  & svm  &s3vm & s4vm & svm/DT  & svm  &s3vm & s4vm & svm/DT\\%& svm  &s3vm & s4vm & svm/DT\\
\hline\noalign{\smallskip}
bci&56.2&-1.0&-1.1&1.8 &62.7&\-1.2&-0.6&0 &64.3&0.9&0&-0.9\\%&65.1&0&0.2&-1.7\\
g241c & 73.7 & 6.6 & 0.4 & 4.2 & 80.7 & 0.9 & 0.2 & 0.1 & 84.2 & 0.9 & 0.8 & 2.1 \\%& 85.0 & 0.2 & 0 & 1.3\\
digit1&90.9&0.6&2.0&1.0&93.7&0.2&0&0.4 &93.9&0&0.1&0.2\\%&96.8&-1.4&-2.3&0.2\\
usps&86.9&-0.2&8.3&1.7&86.9& 0&8.4&1.8&90.4&0&0.1&0.4\\%&91.5&-0.2&0&-0.1\\
heart scale&71.2&0.9&1.3&1.0&78.9&-0.6&0.7&1.9&80.8&0.3&0.2&5.7\\%&85.1&-0.5&0&0\\
vehicle& 65.3&0.2&2.3&0.8&74.2&2.3&3.4&-0.1&77.0&3.0&-0.9&-1.2\\%&79.5&0.3&0.5&-2.7\\
\hline\noalign{\smallskip}
\multicolumn{4}{c|}{}&\multicolumn{3}{|c|}{s3vm}&\multicolumn{3}{|c|}{s4vm}&\multicolumn{3}{|c}{svm/DT}\\
\hline\noalign{\smallskip}
\multicolumn{4}{l|}{Win/Tie/Loss against svm:}&\multicolumn{3}{|c|}{ 12/3/3}&\multicolumn{3}{|c|}{ 13/2/3}&\multicolumn{3}{|c}{ \textbf{14/1/3}} \\
\multicolumn{4}{l|}{Best/Worst performance over all:}&\multicolumn{3}{|c|}{ 5/8}&\multicolumn{3}{|c|}{6/5}&\multicolumn{3}{|c}{\textbf{ 7/4}} \\
\hline
\end{tabular}
\end{center}
\vskip -0.1in
\end{table*}
\setlength{\tabcolsep}{3pt}
\subsection{Comparison with Classical Semi-supervised Wrapper Methods}

This part compares Dual Teaching (DT) with previous semi-supervised wrapper methods, Self-Training (ST) and Co-Training (CT). To show the generalization of the proposed method, Logistical Regression (lr)~\cite{hosmer2004applied}, Support Vector Machine with linear kernel (svm)~\cite{suykens1999least} and Adaboost (ada)~\cite{ratsch2001soft} serve as the base learner separately for every wrapper method. Every specific method is abbreviated to "the name of the base learner/the name of the wrapper method". For instance, "lr/ST" means the base learner is a logistic regression model and the wrapper method is Self-Training. To show whether the wrapper methods can improve the performance of the base learner from unlabeled data, supervised models with the use of labeled data only, which are abbreviated to "name of the supervised model", serve as the baseline approaches. For instance, "lr" represents the logistic regression model trained only from labeled data. This experiment contains all data sets referred above. For every data set, each method works on various numbers of labeled data and the performance on testing set is recorded. The ratio of the number of labeled data to that of all data in training set ranges from 0.1 to 1. When the ratio is equal to 1, the base learner actually is a supervised classifier trained from fully-labeled data.

Fig~\ref{fig:exp} shows the comparison between Dual Teaching and other semi-supervised wrapper methods. Dual Teaching performs much better than Self-Training and Co-Training. Except that the training data are all labeled, there are 324 different cases (3 base learners, 12 data sets and 9 ratios of labeled data). In 235(72.3\%) cases, Dual Teaching makes more accurate predictions than the other methods and only in 9(2.8\%) cases, Dual Teaching is worse than others. In addition, only in 11 cases, Dual Teaching is outperformed by the baseline model using labeled data only. Furthermore, for almost all data sets and different types of supervised models, Dual Teaching can train a base learner from 20\% labeled data and 80\% unlabeled as well as a fully-labeled-data-trained classifier. In summary, Dual Teaching can effectively improve the performance of the base learner from unlabeled data.

%In addition, the situation that Dual Teaching is outperformed by the base learners occurs 15 times far less than.

\subsection{Comparison with Productive Semi-supervised Methods}
The proposed method is also compared with some of productive semi-supervised methods (no wrapper method): S3VM~\cite{Bennett1999Semi} and S4VM~\cite{Li2015Towards} with linear kernel. The base learner of Dual Teaching is set to a SVM with linear kernel too. the ratio of the number of labeled data to that of training data ranges in $\{5\%,\ 10\%,\ 20\%$\}. To validate the safeness of the above semi-supervised methods, a SVM classifier with linear kernel trained from labeled data only serves as the baseline approach.
%The number of labeled data change in this experiment and the ratio of the number of labeled data to that of training data ranges in $\{5\%,\ 10\%,\ 20\%,\ 40\%$\}. SVM with linear kernel trained from labeled data only serve as the baseline approach.

Table~\ref{table:comparison} presents the accuracy of SVM and accuracy improvements of S3VM, S4VM and Dual Teaching on different numbers of labeled data. In the perspective of no matter effectiveness or safeness, Dual Teaching is competitive in this experiment. Different from S3VM and S4VM, Dual Teaching is a wrapper method that can make full use of the supervised learning according to different data sets. For example, as shown in Fig~\ref{fig:exp}, Dual Teaching with adaboost as the base learner performs better than Dual Teaching with svm as the base learner on \emph{bci} and \emph{vehicle}. %This is a special advantage that other semi-supervised methods do not have.

\section{Conclusion and Future Work}
Dual Teaching's good performance relies on its easy assumptions and novel framework. The assumptions are easy to set up in practice. First, the assumptions themselves are not critical. For instance, with the base learner improves, the performance of teachers are usually worse than that in early generations. But the teachers can still match the assumptions just by reducing the recall to keep their precision in a satisfactory level. Then, the assumptions are well quantified so that we can monitor and tune the external teachers in the process of Dual Teaching to match the necessary conditions. Dual Teaching can effectively improve the performance of supervised learner from unlabeled data.

The above characters also improve the safety of Dual Teaching. By tuning the teachers, the base learner in Dual Teaching is able to be not worse than a same supervised model trained from labeled data only. When the recall of each teacher is equal to 0, the re-training set does not change anymore and the base learner is constant. To estimate the performance of teachers precisely, Dual Teaching needs enough labeled data reserved (at least 10\% empirically). Unlike methods which focus on few labeled data training~\cite{Zhou2007Semi}, Dual Teaching does not perform well in the cases where only few labeled data exist.

The easy assumptions originate from the carefully designed framework. Dual Teaching converts training an high performance base learner into training two qualified external classifiers in every generation. The re-training set of the base learner synthesizes the outputs of external classifiers in every generation and boosts the performance of the base learner. Even though the teachers in late generations are poor, only if they do not pollute the re-training set too much, the base learner will not grow worse. For high effectiveness, Dual Teaching sacrifices its efficiency. Dual Teaching re-trains the base learner in every generation and the scale of re-training set increases in the recursive process of Dual Teaching.

In the future, it is worthy to extend Dual Teaching to solve multiple classification problems. Designing a more effective and efficient way to estimate the precision and the recall of auxiliary classifiers is also worthy.

%\bigskip
%\noindent Thank you for reading these instructions carefully. We look forward to receiving your electronic files!
\newpage
\bibliography{example_paper}
\bibliographystyle{aaai}
\end{document}